\documentclass{article}

\usepackage[preprint]{neurips_2026}

\usepackage[utf8]{inputenc} 
\usepackage[T1]{fontenc}    
\usepackage{hyperref}       
\usepackage{url}            
\usepackage{booktabs}       
\usepackage{amsfonts}       
\usepackage{nicefrac}       
\usepackage{microtype}      
\usepackage{xcolor}         
\usepackage{amsmath}
\usepackage{bbm}
\usepackage{graphicx}
\usepackage{multirow}
\usepackage{arydshln}
\usepackage[table]{xcolor}
\usepackage{wrapfig}

\usepackage{algorithm}
\usepackage{algorithmicx}
\usepackage{algpseudocode}
\usepackage{float}

\usepackage[table]{xcolor}
\definecolor{lightblue}{RGB}{173,216,230}

\usepackage{tabularx}
\newcolumntype{C}{>{\centering\arraybackslash}X}

\newtheorem{proposition}{Proposition}

\usepackage{graphicx}
\usepackage{wrapfig}

\title{Decoupled Guidance Diffusion for Adaptive Offline Safe Reinforcement Learning}

\author{%
  Rufeng Chen$^{1}$ \quad
  Zhaofan Zhang$^{1}$ \quad
  Zhejian Yang$^{2}$ \quad
  Hechang Chen$^{2}$ \quad
  Sihong Xie$^{1}$ \\[0.5em]
  $^{1}$The Hong Kong University of Science and Technology (Guangzhou) \\
  $^{2}$Jilin University \\
}

\begin{document}

\maketitle

\begin{abstract}
Offline safe reinforcement learning often requires policies to adapt at deployment time to safety budgets that vary across episodes or change within a single episode.
While diffusion-based planners enable flexible trajectory generation, existing guidance schemes often treat reward improvement and constraint satisfaction as competing gradient objectives, which can lead to unreliable safety compliance under cost limits.
We reinterpret adaptive safe trajectory generation as sampling from a constrained trajectory distribution, where the budget restricts the trajectory region, and reward shapes preferences within that region.
This perspective motivates \textbf{Safe Decoupled Guidance Diffusion (SDGD)}, which conditions classifier-free guidance on the cost limit to bias sampling toward trajectories satisfying the specified limit, while using reward-gradient guidance to refine trajectories for higher return.
Because direct reward guidance can increase return while also steering samples toward trajectories with higher cumulative cost, we introduce Feasible Trajectory Relabeling (FTR) to reshape reward targets and discourage such directions.
We further provide a first-order sampling-time analysis showing that FTR suppresses reward-induced cost drift under a prefix-restorative alignment condition. 
Extensive evaluations on the DSRL benchmark show that SDGD achieves the strongest safety compliance among baselines, satisfying the constraint on 94.7\% of tasks (36/38), while obtaining the highest reward among safe methods on 21 tasks.

\end{abstract}

\section{Introduction}

Offline safe reinforcement learning aims to learn policies from pre-collected datasets that optimize task performance while satisfying safety constraints \cite{brahmanage2026beyond, ma2025constraint}.
In deployment, however, safety budgets are often not fixed once and for all: they may vary across tasks, users, or operating conditions, and may even change within a single episode.
This creates a challenging adaptive safety problem, where a planner must respond to deployment-time cost limits without collecting new data or retraining the policy.
Such requirements are common in safety-critical domains such as robotics \cite{gao2024coohoi, hoang2025geometryaware, hung2025victor} and autonomous driving \cite{duan2024learning,peng2024navigating, zheng2025diffusionbased, he2024everyday}, but existing offline safe RL methods are typically designed for fixed or weakly varying constraints.
Designing offline planners that can reliably adapt to changing cost limits therefore remains a central challenge.

\begin{figure}
\begin{center}
\vskip -0.1in
\centerline{\includegraphics[width=\textwidth]{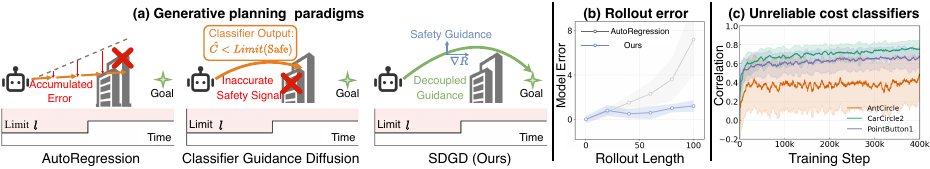}}
\vskip -0.1in
\caption{
Motivation for SDGD. 
(a) Autoregressive planning suffers from accumulated rollout errors, while classifier-guided diffusion can be misled by inaccurate safety signals. 
SDGD decouples safety and reward guidance using CFG for safety enforcement and $\nabla R$ for reward optimization. 
(b) SDGD reduces rollout error compared with autoregressive generation. 
(c) Cost classifiers show low and unstable correlation with true cumulative cost, indicating unreliable safety guidance.
}

\label{fig:motivation}
\end{center}
\vskip -0.3in
\end{figure}

Existing approaches to adaptive safe offline reinforcement learning commonly use constraint-conditioned policies \cite{guo2025constraint, chemingui2025constraint} or autoregressive trajectory models \cite{su2026boundarytoregion}. 
Although these methods can condition decisions on deployment-time cost limits, their step-wise or sequential generation process makes trajectory-level cost control difficult. 
Because safety constraints depend on cumulative costs over long horizons, small local prediction errors can compound and lead to large deviations in trajectory-level cost \cite{park2025scalable, liu2025efficient}. 
Figure~\ref{fig:motivation}(a) illustrates this error propagation, and Figure~\ref{fig:motivation}(b) provides a controlled diagnostic showing that one-step rollout generation accumulates error with horizon length, whereas SDGD generates trajectory segments jointly and yields smaller accumulated error. 
These observations suggest that sequential generation can be vulnerable to constraint violations under cost limits.


Diffusion-based planners offer a promising alternative by generating trajectory segments jointly rather than autoregressively, which can reduce the error accumulation of sequential generation \cite{zhang2025constrained, xiao2023safediffuser}. 
However, existing safe diffusion planners often rely on classifier guidance or gradient switching, where cost predictions on intermediate noisy trajectories are used to decide how strongly cost guidance should override reward guidance \cite{lin2023safe}. 
This requires the cost classifier to provide reliable safety signals throughout reverse diffusion. 
Empirically, this assumption can be fragile: Figure~\ref{fig:motivation}(c) shows that the Pearson correlation between predicted and true cumulative cost is low and unstable across multiple DSRL tasks. 
As a result, classifier-guided or switching-based methods can apply cost gradients at the wrong time, leading to unreliable constraint satisfaction under deployment-time cost limits.


To address this limitation, we reinterpret adaptive safe trajectory generation as sampling from a constrained trajectory distribution. 
In this view, the cost limit restricts the region of trajectories that should be sampled, while the reward only shapes preferences among trajectories within that region. 
This reveals an asymmetry between safety and reward: the former is a restriction on the trajectory distribution, whereas the latter is a preference for improving return. 
Therefore, constraint satisfaction and reward improvement should not be treated as interchangeable gradient objectives. 
Motivated by this observation, we propose to decouple the two roles by using classifier-free guidance to incorporate the cost limit and reward-gradient guidance to improve return.


Building on this perspective, we propose \textit{Safe Decoupled Guidance Diffusion} (SDGD), a diffusion-based planner for adaptive safe offline reinforcement learning. 
SDGD conditions classifier-free guidance on the cost limit to bias sampling toward trajectories satisfying the specified limit, while using reward-gradient guidance to improve return. 
Because direct reward guidance can improve return but may inadvertently increase cumulative cost, we introduce \textit{Feasible Trajectory Relabeling (FTR)} to reshape reward targets and reduce such cost increases.
We further provide a first-order sampling-time analysis showing that FTR suppresses reward-induced cost drift under a prefix-restorative alignment condition. 
Across the DSRL benchmark \cite{liu2023datasets}, SDGD adapts to changing cost limits without retraining, satisfies the normalized cost constraint on 94.7\% of tasks (36/38), and achieves the best reward among methods satisfying the constraint on 21 tasks.


\section{Related work}

\textbf{Offline Safe RL.}
Offline safe RL studies how to optimize task performance while satisfying safety constraints from fixed offline datasets. 
Early methods mainly considered fixed safety constraints while addressing distribution shift, using Lagrangian methods (BCQ-Lag) \cite{stooke2020responsive}, distribution correction (COptiDICE) \cite{lee2022coptidice}, penalty-based objectives (CPQ) \cite{xu2022constraints}, and feasibility-guided learning (FISOR) \cite{zheng2024safe}. 
SDQC \cite{yang2025qsupervised} further studies representation learning to decouple safety-relevant and reward-relevant features. 
More recent work considers adaptation to different cost limits without retraining, including policy-conditioning methods such as CCAC \cite{guo2025constraint}, policy-switching methods such as CAPS \cite{chemingui2025constraint}, constraint-conditioned value learning such as C2IQL \cite{liu2025ciql}, and generative models such as CDT \cite{liu2023constrained}, B2R \cite{su2026boundarytoregion} and TREBI \cite{lin2023safe}. 
In contrast, SDGD treats the cost limit as a restriction on the trajectory distribution and separates this role from reward-guided refinement during diffusion sampling.


\textbf{Trajectory Optimization with Planning.}
Trajectory planning has long been studied in model-based decision making \cite{nagabandi2020deep, hamrick2020role} and has recently benefited from sequence and generative models. 
Transformer-based planners \cite{chen2021decision, huang2024context} generate actions or trajectory tokens autoregressively, while diffusion planners such as Diffuser \cite{janner2022planning}, Decision Diffuser \cite{ajay2022conditional}, M2Diffuser \cite{yan2024m2diffuser}, and Diffusion Policy \cite{chi2023diffusion} denoise trajectory segments jointly and enable flexible guidance. 
Latent diffusion methods \cite{venkatraman2023reasoning, li2023efficient} further improve planning through compact trajectory representations. 
Unlike prior guided planners that treat reward, task, or constraint signals as conditioning variables or gradient objectives, SDGD separates the roles of cost limits and rewards: the cost limit shapes the sampled trajectory distribution, while reward guidance improves return.


\begin{figure}
\begin{center}
\centerline{\includegraphics[width=\textwidth]{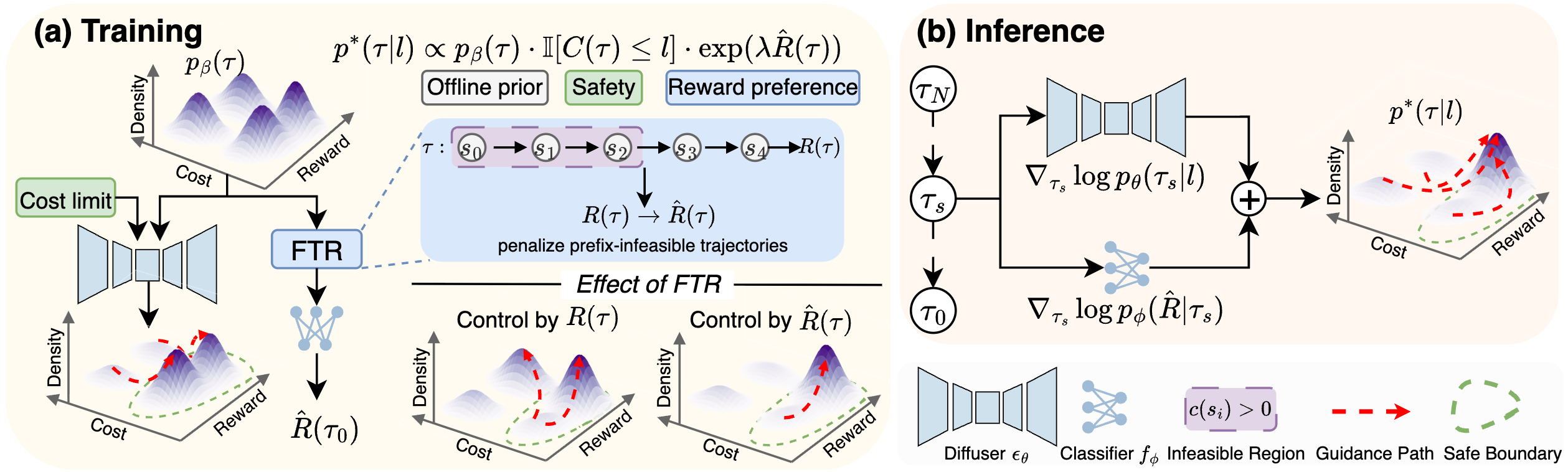}}
\label{fig:overview}
\caption{
Overview of Safe Decoupled Guidance Diffusion (SDGD).
(a) SDGD learns a cost limit ($l$) conditioned diffuser to model the safe trajectory distribution from offline data, while an FTR-trained reward classifier predicts rewards.
Feasible Trajectory Relabeling (FTR) modifies reward single by replacing the raw reward $R(\tau)$ with a new reward target $\hat{R}(\tau)$, thereby reshaping reward guidance away from infeasible regions and reducing reward-induced cost drift.
(b) SDGD decouples safety and performance: classifier-free guidance enforces cost limits, while reward-gradient guidance refines samples toward high-return trajectories within the safe region.
}

\label{fig:method}
\end{center}
\vskip -0.3in
\end{figure}

\section{Preliminaries}
\subsection{Safe Reinforcement Learning}
Safe reinforcement learning is commonly formulated as a constrained Markov decision process (CMDP) \cite{ames2019control}.
A CMDP is defined by the tuple $\mathcal{M} = (\mathcal{S},\mathcal{A},{T},r,c,l,\gamma)$, where $\mathcal{S}$ represents the state space; $\mathcal{A}$ denotes the action space; $T = \mathcal{S} \times \mathcal{A} \to \Delta(\mathcal{S})$ describes the transition dynamics; $r = \mathcal{S}\times\mathcal{A} \to [R_{\text{min}}, R_{\text{max}}] $ is the state-action reward function; $ c = \mathcal{S}\times\mathcal{A}\to[0, C_{\text{max}}]$ is the state-action cost function; $l$ is a cost limit and $\gamma \in (0,1]$ is the discount factor. 
The finite trajectory $\tau = \{s_i, a_i,..., s_{i+L}, a_{i+L}\}$, where $L$ is the maximum episode length. 
The total discounted reward and cost for a trajectory $\tau$ are defined as $R(\tau) = \sum_{t=i}^{i+L} \gamma^{t-i}r(s_t,a_t)$ and $C(\tau) = \sum_{t=i}^{i+L} \gamma^{t-i}c(s_t,a_t)$, respectively.
The core objective combines reward maximization with time-varying safety constraints:
\begin{equation}
\max_\pi\mathbb{E}_{\tau\sim\pi}[R(\tau)]\quad
\text{s.t.} \; \mathbb{E}_{\tau\sim\pi}[C_{k}(\tau)] \leq l_k, \; \forall k \in {0,...,K}.
\label{Eq:general_objective}
\end{equation}
Here the $\mathbb{E}_{\tau\sim\pi}[\cdot]$ denotes the expectation over trajectories $\tau$ generated by the policy $\pi$. 
Each $C_{k}(\tau) = \sum_{t=k}^{k+T_k}\gamma^{t-k}c(s_t,a_t)$ is the cost of sub-trajectories with predefined horizon $T_k$, constrained by its corresponding limit $l_k$.
This general objective naturally covers fixed limits ($K=0$) and time-varying limits ($K>0$) through segmented cost limits.

\subsection{Diffusion Models}
Diffusion models learn a data distribution by defining a forward noising process and training a neural network to reverse it. 
Starting from a clean sample $x_0 \sim p_{\mathrm{data}}$, the forward process gradually perturbs the data with Gaussian noise over $N$ diffusion steps. 
We use $s \in \{1,\ldots,N\}$ to denote the diffusion time index, reserving $t$ for environment time in trajectories throughout the paper. 
Under this process, the noisy sample at diffusion step $s$ can be written directly as:
\begin{equation}
q(x_s\mid x_0)
=
\mathcal{N}\left(
x_s;
\sqrt{\bar{\alpha}_s}x_0,
(1-\bar{\alpha}_s)I
\right),
\end{equation}
where $\bar{\alpha}_s=\prod_{i=1}^{s}\alpha_i$, and $\{\alpha_s\}_{s=1}^{N}$ is a predefined variance schedule. 
A neural network $\epsilon_\theta(x_s,s)$ is trained to predict the injected noise $\epsilon$ by minimizing
\begin{equation}
\mathcal{L}(\theta)
=
\mathbb{E}_{s,x_0,\epsilon}
\left[
\left\|
\epsilon
-
\epsilon_\theta
\left(
\sqrt{\bar{\alpha}_s}x_0
+
\sqrt{1-\bar{\alpha}_s}\epsilon,
s
\right)
\right\|^2
\right].
\end{equation}

\paragraph{Classifier Guidance.}
Classifier guidance uses gradients of a separately trained classifier $p_\phi(y\mid x_s)$ to adjust the reverse diffusion direction:
\begin{equation}
\widehat{\epsilon}_\theta(x_s,s,y)
=
\epsilon_\theta(x_s,s)
+
\lambda \Sigma_s \nabla_{x_s}\log p_\phi(y\mid x_s),
\end{equation}
where $\lambda$ controls the classifier-guidance strength.

\paragraph{Classifier-Free Guidance.}
Classifier-free guidance jointly learns conditional and unconditional denoising models $\epsilon_\theta(x_s,s,y)$ and $\epsilon_\theta(x_s,s,\emptyset)$. 
Conditional sampling is achieved by
\begin{equation}
\widehat{\epsilon}_\theta(x_s,s,y)
=
w\epsilon_\theta(x_s,s,y)
+
(1-w)\epsilon_\theta(x_s,s,\emptyset),
\end{equation}
where $w$ controls the conditioning strength.

\section{Methods}\label{Methods}

\subsection{Constrained Trajectory Posterior Formulation}\label{Constrained Trajectory Posterior Formulation}
We consider offline safe trajectory generation under a cost limit $l$.
Our goal is to learn a trajectory generator that, for any specified $l$, samples trajectories that remain within the cost limit while favoring high reward.
Rather than learning a policy tied to a fixed constraint level, we formulate this goal as sampling from a cost-conditioned target distribution $p^*(\tau|l)$.
Let $p_\beta(\tau)$ denote the trajectory prior induced by the offline dataset.
The target distribution over trajectories can be written as:
\begin{equation}
    \label{objective}
    p^*(\tau|l) \propto p_\beta(\tau) \cdot \mathbb{I}[C(\tau) \le l] \cdot \exp(\lambda R(\tau))
\end{equation}
where $R(\tau)$ and $C(\tau)$ denote trajectory return and cumulative cost, respectively, and $\lambda $ controls the reward preference strength.
This formulation highlights a structural asymmetry between safety and performance.
The dataset prior \(p_\beta(\tau)\) anchors sampling to the offline data manifold; the indicator \(\mathbb{I}[C(\tau)\le l]\) restricts the target support to trajectories satisfying the cost limit; and the exponential reward term reshapes the relative likelihood of trajectories within that support.
Although the indicator term does not define a smooth potential, a formal score-level decomposition illustrates the different roles of the cost-limit and reward terms:
\begin{equation}
    \label{asymmetry}
    \nabla_\tau \log p^*(\tau|l) = \nabla_\tau \log p_\beta(\tau) + \nabla_\tau \log \mathbb{I}[C(\tau) \le l] + \lambda  \nabla_\tau R(\tau)
\end{equation}
The reward component naturally admits gradient-based refinement. 
In contrast, the cost-limit component represents a non-differentiable support restriction and is not well suited to direct local gradient shaping.
This distinction motivates a decoupled guidance architecture, where the cost limit shapes the sampled trajectory distribution and reward guidance refines trajectories for higher return.

\subsection{Safe Decoupled Guidance Diffusion}
\label{Safe Decoupled Guidance Diffusion}
SDGD instantiates the target distribution in Eq.~(\ref{objective}) by assigning the two non-prior terms to different guidance mechanisms.
The cost limit restriction is incorporated through a cost-conditioned diffusion model, while reward optimization is introduced as local gradient refinement during reverse sampling.

\paragraph{Cost-conditioned diffusion.}
For each trajectory in the offline dataset, we compute its cumulative cost and use it to construct cost-limit conditioning.
The conditional diffusion model estimates the score of noisy trajectories associated with clean trajectories satisfying the specified limit: $s_\theta(\tau_s,s,l) \approx \nabla_{\tau_s}\log p_s(\tau_s \mid C(\tau_0)\le l),$ where \(p_s\) denotes the distribution obtained by applying \(s\) forward-noising steps to trajectories \(\tau_0\) with \(C(\tau_0)\le l\).
At deployment time, the desired cost limit \(l\) can be specified directly, including limits that vary across episodes or across planning steps.
To strengthen conditioning, SDGD applies classifier-free guidance:
\begin{equation}
s_{\mathrm{safe}}(\tau_s,s,l)
=
(1+w)s_\theta(\tau_s,s,l)
-
w s_\theta(\tau_s,s,\emptyset),
\end{equation}
where \(s_\theta(\tau_s,s,\emptyset)\) is the unconditional score and \(w\) controls the conditioning strength.
This avoids noisy cost classifiers and incorporates the cost limit through the denoising model itself.

\paragraph{Reward-gradient refinement.}
Given the cost-conditioned score, SDGD further biases sampling toward high-return trajectories using a learned reward predictor \(R_\phi\).
Following the reward-tilted posterior in Eq.~(\ref{objective}), the sampling score is augmented as:
\begin{equation}
\label{eq:score_sdgd}
s_{\mathrm{SDGD}}(\tau_s,s,l)
=
s_{\mathrm{safe}}(\tau_s,s,l)
+
\lambda \nabla_{\tau_s} R_\phi(\tau_s),
\end{equation}
where \(\lambda\) is the reward guidance scale.
The first term biases denoising toward the cost-conditioned trajectory distribution, while the second term improves return through local reward-gradient refinement.

This decoupled sampler differs from classifier-guided safety methods in a key aspect: constraint satisfaction does not rely on differentiating a cost predictor on noisy trajectories.
Instead, the cost limit enters as a generative condition through classifier-free guidance, and classifier-based gradients are reserved for reward refinement.
As a result, the same trained model can adapt to different or dynamically changing cost limits without retraining.

\subsection{Feasibility-Consistent Reward Guidance}
\label{Feasibility Consistent Reward Guidance}
Although reward-gradient guidance improves return, applying it directly to raw cumulative rewards can undermine the safety-conditioned sampler in Eq.~(\ref{eq:score_sdgd}). 
The issue is not that the cost-conditioned diffusion model fails to represent safe trajectories, but that the reward gradient acts as an additional sampling-time perturbation. 
In offline datasets, high-return trajectories may also incur high costs, especially when task completion and constraint violation are correlated. 
As a result, a reward predictor trained on raw returns can assign large gradients to trajectory features that are predictive of both high reward and high cost. 
When $\langle \nabla_\tau R(\tau), \nabla_\tau C(\tau) \rangle > 0$, following the reward gradient increases not only reward but also cumulative cost, producing a reward-induced cost drift.

\paragraph{Feasible Trajectory Relabeling.}
To make reward guidance compatible with the safety-conditioned sampler, we introduce \textit{Feasible Trajectory Relabeling} (FTR). 
FTR modifies the target used to train the reward predictor, while leaving the diffusion model and the cost limit conditioning unchanged.

The key idea is to distinguish cumulative safety violation from prefix-level execution risk. 
In receding-horizon planning, only the first few steps of a generated trajectory are executed before replanning. 
Costs incurred in this executable prefix immediately affect the environment and cannot be revised by future planning, whereas costs appearing later may still be avoided or regulated by subsequent cost-conditioned denoising and replanning. 
Therefore, we treat trajectories that incur cost in the executable prefix as prefix-infeasible for the purpose of reward relabeling.
Let \(f\) denote the executable prefix length. 
We define the prefix-infeasibility indicator as $h_f(\tau)=\mathbf{1}\left[\sum_{t=0}^{f-1} c(s_t,a_t)>0\right],$
and construct the FTR reward target: $\widehat{R}(\tau)=R(\tau)+r_{\mathrm{us}} h_f(\tau), $ where $r_{us} < 0.$

The reward predictor \(R_\phi\) is then trained to regress \(\widehat{R}(\tau)\) instead of the raw return \(R(\tau)\). 
Thus, FTR penalizes trajectories whose high return depends on prefix-infeasible behavior, without suppressing all high-return trajectories indiscriminately.



\paragraph{How FTR reshapes reward guidance.}
FTR is not merely a scalar penalty applied to prefix-infeasible trajectories.
Its purpose is to change the reward-gradient field learned by $R_\phi$. 
As illustrated in Figure~\ref{fig:overview}(a), using the raw return \(R(\tau)\) can induce a learned preference distribution in which high return and high cost co-occur.
By training the reward predictor on the relabeled target \(\widehat{R}(\tau)\), FTR reduces the coupling between high return and high cost and biases reward guidance away from return improvements that also increase cost.

For analysis, we use a differentiable surrogate of \(h_f\) so that \(\nabla_\tau h_f\) is well defined.
Under this surrogate, the idealized relabeled reward gradient can be written as
\begin{equation}
\nabla_{\tau_s} \widehat{R}_\phi(\tau_s)
=
\nabla_{\tau_s} R_\phi(\tau_s)
+
r_{\mathrm{us}}\nabla_{\tau_s}h_f(\tau_s).
\end{equation}
Since \(r_{\mathrm{us}}<0\), the second term introduces a corrective direction away from prefix-infeasible regions.

To quantify this effect, we compare three reverse trajectories coupled from the same noisy initialization and reverse-sampling randomness:
\(\tau^C\), \(\tau^R\), and \(\tau^{\widehat{R}}\), corresponding to the cost-conditioned, raw reward-guided, and FTR reward-guided samplers, respectively.
We measure the final cost drift introduced by reward guidance relative to the cost-conditioned sampler as
$\Delta C_R
=
C(\tau_0^{R})-C(\tau_0^{C}),
$ and $
\Delta C_{\widehat{R}}
=
C(\tau_0^{\widehat{R}})-C(\tau_0^{C}).$
Under a first-order perturbation approximation of the full reverse denoising process, their difference satisfies
\begin{equation}
\Delta C_{\widehat{R}}-\Delta C_R
\approx
\eta r_{\mathrm{us}} A_f(N)
+
\epsilon_\Delta ,
\end{equation}
where \(\eta>0\) denotes the effective perturbation scale induced by reward guidance over the reverse denoising process, and \(\epsilon_\Delta\) denotes higher-order approximation error.
\begin{equation}
A_f(N)
=
\int_0^N
\kappa_s
\left\langle
\nabla_\tau C(\tau_s^C),
\nabla_\tau h_f(\tau_s^C)
\right\rangle ds .
\end{equation}
Here, \(s\) is the integration variable along reverse diffusion time, and \(\kappa_s\) collects the time-dependent reverse-sampling factors at time \(s\).
We evaluate the alignment along the cost-conditioned reference path \(\tau_s^C\), because both raw and FTR reward guidance are treated as first-order perturbations around the same cost-conditioned sampler.

Thus, \(A_f(N)\) measures the accumulated alignment between the cumulative cost gradient and the prefix-infeasibility gradient along the cost-conditioned reverse trajectory.
When \(A_f(N)>0\), moving toward larger values of the prefix-infeasibility surrogate also increases cumulative cost.
Since \(r_{\mathrm{us}}<0\), the FTR correction contributes a negative first-order term, \(\eta r_{\mathrm{us}}A_f(N)\), to the cost drift relative to raw reward guidance.
When this alignment is consistently positive, we denote by \(\Gamma_f>0\) a lower bound such that \(A_f(N)\geq \Gamma_f\) with high probability.
This quantity captures the minimum effective strength of the FTR correction: larger \(\Gamma_f\) means that moving away from prefix-infeasible directions more directly reduces cumulative cost.



\begin{proposition}[Relative suppression of reward-induced cost drift]
\label{prop:reward_induced_cost_drift}
For the coupled reverse trajectories \(\tau^C\), \(\tau^R\), and \(\tau^{\widehat{R}}\), suppose that \(A_f(N)\geq \Gamma_f>0\) with probability at least \(1-\delta\), and that the first-order residual difference satisfies \(|\epsilon_\Delta|\leq \bar{\epsilon}_\Delta\). 
Then, with probability at least \(1-\delta\),
\begin{equation}
C(\tau_0^{\widehat{R}})
\leq
C(\tau_0^{R})
-
\eta |r_{\mathrm{us}}|\Gamma_f
+
\bar{\epsilon}_\Delta ,
\end{equation}
where \(\eta>0\) is the effective perturbation scale induced by reward guidance.
Consequently, if \(\eta |r_{\mathrm{us}}|\Gamma_f>\bar{\epsilon}_\Delta\), the FTR-guided sampler induces a smaller cost deviation than raw reward guidance relative to the cost-conditioned sampler.
\end{proposition}

Proposition~\ref{prop:reward_induced_cost_drift} shows that FTR reduces the cost increase introduced by reward guidance when the infeasibility direction is positively aligned with cost. 
Importantly, FTR does not replace the cost-conditioned diffusion model, nor does it introduce a cost classifier. 
Instead, it makes reward guidance less aligned with cost-increasing directions, allowing reward refinement to remain compatible with the safety support learned by classifier-free guidance.

\paragraph{Practical choice of the penalty.}
The penalty \(r_{\mathrm{us}}\) is chosen to create label-level separation between prefix-infeasible and prefix-feasible trajectories.
Specifically, after relabeling, even the largest possible return of a prefix-infeasible trajectory should be lower than the smallest possible return of a prefix-feasible trajectory.
Let \(r_{\min}\) and \(r_{\max}\) denote the minimum and maximum per-step rewards observed in the dataset, and let \(L\) be the planning horizon.
A worst-case upper bound on the return of any prefix-infeasible trajectory is $R^*=\frac{r_{\max}(1-\gamma^{L})}{1-\gamma},$ while a worst-case lower bound on the return of a prefix-feasible trajectory is $R^o=\frac{r_{\min}(1-\gamma^{L})}{1-\gamma}.$
Requiring \(R^*+r_{\mathrm{us}}<R^o\) gives $r_{\mathrm{us}} < \frac{(r_{\min}-r_{\max})(1-\gamma^{L})}{1-\gamma}.$
This choice makes prefix-infeasible trajectories unattractive to the reward predictor even when they have high raw return. In practice, we compute \(r_{\min}\) and \(r_{\max}\) from the offline dataset and use this bound as a conservative guideline.


\begin{wrapfigure}{b}{0.55\textwidth}
\vskip -0.2in
\centering
\includegraphics[width=0.55\textwidth]{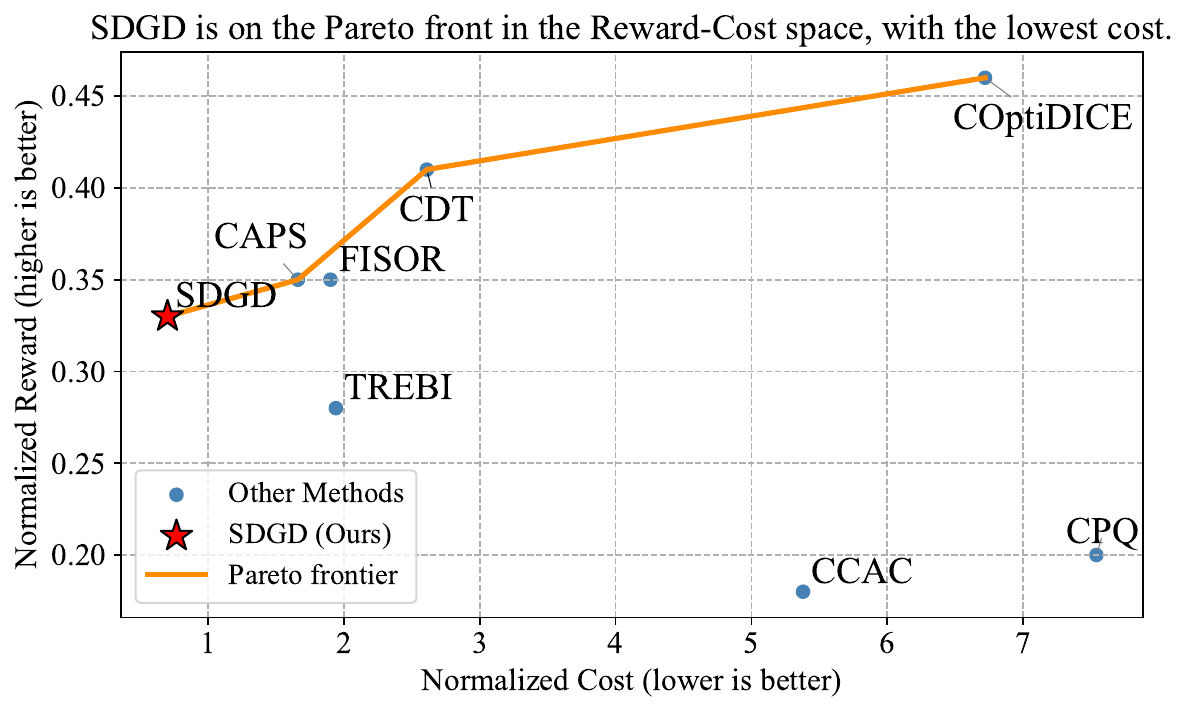}
\vskip -0.1in
\caption{
    Aggregate reward-cost performance, averaged over all 38 DSRL tasks. 
    Safety is defined as a normalized cost $\leq 1$. SDGD is the only method to achieve an average cost within this safe region.
}
\label{fig: pareto}
\vspace{-0.1in}
\end{wrapfigure}

\section{Experiments}
\label{Experiments}

\begin{table*}[t]
\vskip -0.1in
\caption{Normalized DSRL \cite{liu2023datasets} benchmark results. $\uparrow$ means the higher the better. $\downarrow$ means the lower the better. Each value is averaged over 20 evaluation episodes and 3 seeds. \textcolor{gray}{Gray}: Unsafe agents. \textbf{Bold}: Safe agents whose normalized cost is smaller than 1.  \textbf{\textcolor{blue}{Blue}}: Safe agents with the highest reward.}
\vskip -0.1in
\label{tab: main_result1}
\setlength{\tabcolsep}{3pt} 
\scriptsize
\begin{center}
\resizebox{\textwidth}{!}{
\begin{tabular}{c cc cc cc cc cc cc cc cc} 
\toprule
\multirow{2}{*}{Task}& \multicolumn{2}{c}{COptiDICE} & \multicolumn{2}{c}{CPQ} & \multicolumn{2}{c}{CDT} & \multicolumn{2}{c}{TREBI} & \multicolumn{2}{c}{FISOR} & \multicolumn{2}{c}{CAPS}  & \multicolumn{2}{c}{CCAC} & \multicolumn{2}{c}{SDGD \small(ours)}  \\
\cline{2-3} \cline{4-5} \cline{6-7} \cline{8-9} \cline{10-11} \cline{12-13} \cline{14-15} \cline{16-17}
\addlinespace[2pt]
 & reward$\uparrow$ & cost$\downarrow$ & reward$\uparrow$ & cost$\downarrow$ & reward$\uparrow$ & cost$\downarrow$ & reward$\uparrow$ & cost$\downarrow$ & reward$\uparrow$ & cost$\downarrow$ & reward$\uparrow$ & cost$\downarrow$ & reward$\uparrow$ & cost$\downarrow$ & reward$\uparrow$ & cost$\downarrow$ \\
\midrule

CarButton1 &
\textcolor{gray}{-0.16} & \textcolor{gray}{4.63} &  
\textcolor{gray}{0.22} & \textcolor{gray}{40.06} & 
\textcolor{gray}{0.17} & \textcolor{gray}{7.05} & 
\textcolor{gray}{0.07} & \textcolor{gray}{3.75} & 
\textbf{-0.19} & \textbf{0.85} & 
\textbf{\textcolor{blue}{-0.02}}  & \textbf{\textcolor{blue}{ 0.73 }}& 
\textcolor{gray}{0.47} & \textcolor{gray}{33.25} &
\textbf{{-0.04}} & \textbf{{0.63}} \\

CarButton2 &  
\textcolor{gray}{-0.17} & \textcolor{gray}{3.40} & 
\textcolor{gray}{0.08} & \textcolor{gray}{19.03} & 
\textcolor{gray}{0.23} & \textcolor{gray}{12.87} & 
\textbf{-0.03} & \textbf{0.97} & 
\textbf{\textcolor{blue}{0.00}} & \textbf{\textcolor{blue}{0.25}} & 
\textbf{-0.09} & \textbf{0.60} & 
\textcolor{gray}{0.49} & \textcolor{gray}{35.48} &
\textbf{-0.05} & \textbf{0.62} \\

CarCircle1 & 
\textcolor{gray}{0.70} & \textcolor{gray}{17.69}  & 
\textcolor{gray}{0.22} & \textcolor{gray}{17.40} &
\textcolor{gray}{0.48} & \textcolor{gray}{6.91} & 
\textbf{\textcolor{blue}{0.14}} & \textbf{\textcolor{blue}{0.00}} & 
\textcolor{gray}{0.68} & \textcolor{gray}{11.48} &
\textcolor{gray}{0.51} & \textcolor{gray}{5.97} & 
\textcolor{gray}{-0.46} & \textcolor{gray}{4.00} & 
\textbf{\textcolor{blue}{0.14}} & \textbf{\textcolor{blue}{0.84}} \\

CarCircle2 &
\textcolor{gray}{0.78} & \textcolor{gray}{26.56}  &
\textbf{\textcolor{blue}{0.55}} & \textbf{\textcolor{blue}{0.36}} & 
\textcolor{gray}{0.56} & \textcolor{gray}{11.92} & 
\textbf{0.22} & \textbf{0.00} &
\textcolor{gray}{0.58} &\textcolor{gray}{8.32} &
\textcolor{gray}{0.50} & \textcolor{gray}{3.83} & 
\textcolor{gray}{-0.22} & \textcolor{gray}{20.09} & 
\textbf{0.33} & \textbf{0.15} \\

CarGoal1 & 
\textcolor{gray}{0.43} & \textcolor{gray}{2.81} &  
\textcolor{gray}{0.33} & \textcolor{gray}{4.93} & 
\textcolor{gray}{0.60} & \textcolor{gray}{3.15} & 
\textcolor{gray}{0.41} & \textcolor{gray}{1.16} & 
\textcolor{gray}{0.47} & \textcolor{gray}{1.31} & 
\textcolor{gray}{0.33} & \textcolor{gray}{1.51} & 
\textcolor{gray}{0.84} & \textcolor{gray}{5.62} & 
\textbf{\textcolor{blue}{0.29}} & \textbf{\textcolor{blue}{0.76}} \\

CarGoal2 &  
\textcolor{gray}{0.19} & \textcolor{gray}{2.83} & 
\textcolor{gray}{0.10} & \textcolor{gray}{6.31} &
\textcolor{gray}{0.45} & \textcolor{gray}{6.05} & 
\textcolor{gray}{0.13} & \textcolor{gray}{1.16} & 
\textbf{0.04} & \textbf{0.87} & 
\textcolor{gray}{0.10} & \textcolor{gray}{2.14} &  
\textcolor{gray}{0.94} & \textcolor{gray}{16.97} & 
\textbf{\textcolor{blue}{0.07}} &  \textbf{\textcolor{blue}{0.98}} \\

CarPush1 & 
\textcolor{gray}{0.21} & \textcolor{gray}{1.28} & 
\textbf{0.08} & \textbf{0.77}&
\textcolor{gray}{0.27} & \textcolor{gray}{2.12} & 
\textcolor{gray}{0.26} & \textcolor{gray}{1.03} &
\textcolor{gray}{0.27} & \textcolor{gray}{1.75} & 
\textbf{0.16} & \textbf{0.51} & 
\textcolor{gray}{0.36} & \textcolor{gray}{6.58} & 
\textbf{\textcolor{blue}{0.21}} & \textbf{\textcolor{blue}{0.92}} \\

CarPush2 &
\textcolor{gray}{0.10} & \textcolor{gray}{4.55} & 
\textcolor{gray}{-0.03} & \textcolor{gray}{10.00} & 
\textcolor{gray}{0.16} & \textcolor{gray}{4.60} & 
\textcolor{gray}{0.12} & \textcolor{gray}{2.65} & 
\textbf{\textcolor{blue}{0.20}} & \textbf{\textcolor{blue}{0.96}} & 
\textcolor{gray}{0.10} & \textcolor{gray}{1.89} & 
\textcolor{gray}{-0.07} & \textcolor{gray}{19.62} & 
\textbf{0.04} & \textbf{0.70} \\

\cdashline{1-17}
\addlinespace[3pt]
PointButton1 & 
\textcolor{gray}{0.09} & \textcolor{gray}{3.34} & 
\textcolor{gray}{0.46} & \textcolor{gray}{11.88}  &  
\textcolor{gray}{0.06} & \textcolor{gray}{1.53} &
\textcolor{gray}{0.07} & \textcolor{gray}{3.23} &
\textcolor{gray}{0.06} & \textcolor{gray}{1.24} &
\textbf{-0.04} & \textbf{0.60} &
\textcolor{gray}{-0.48} & \textcolor{gray}{3.21} &
\textbf{\textcolor{blue}{0.02}} & \textbf{\textcolor{blue}{0.81}} \\

PointButton2 &
\textcolor{gray}{0.10} & \textcolor{gray}{2.77} & 
\textcolor{gray}{0.52} & \textcolor{gray}{15.04} &  
\textcolor{gray}{0.25} & \textcolor{gray}{7.18} & 
\textbf{\textcolor{blue}{0.08}} & \textbf{\textcolor{blue}{0.60}} & 
\textcolor{gray}{0.12} & \textcolor{gray}{1.13} &
\textcolor{gray}{0.13} & \textcolor{gray}{4.96} &
\textcolor{gray}{-0.47} & \textcolor{gray}{1.96} &
\textbf{{0.05}} & \textbf{{0.92}} \\

PointCircle1 & 
\textcolor{gray}{0.85} & \textcolor{gray}{18.08} &  
\textcolor{gray}{-0.26} & \textcolor{gray}{3.20} &  
\textbf{\textcolor{blue}{0.53}} & \textbf{\textcolor{blue}{0.42}} & 
\textbf{0.41} & \textbf{0.00} & 
\textcolor{gray}{0.27} & \textcolor{gray}{15.95} &  
\textcolor{gray}{0.19} & \textcolor{gray}{1.74} & 
\textcolor{gray}{-0.52} & \textcolor{gray}{10.13} &
\textcolor{gray}{0.31} & \textcolor{gray}{2.74} \\

PointCircle2 & 
\textcolor{gray}{0.86} & \textcolor{gray}{28.66} & 
\textcolor{gray}{0.10} & \textcolor{gray}{11.50} &  
\textbf{0.44} & \textbf{0.90} & 
\textcolor{gray}{0.40} & \textcolor{gray}{6.62} & 
\textcolor{gray}{0.72} & \textcolor{gray}{14.96} & 
\textbf{0.38} & \textbf{0.00} & 
\textcolor{gray}{0.50} & \textcolor{gray}{2.68} &
\textbf{\textcolor{blue}{0.49}} & \textbf{\textcolor{blue}{0.56}} \\

PointGoal1 & 
\textcolor{gray}{0.50} & \textcolor{gray}{5.17} & 
\textcolor{gray}{0.51} & \textcolor{gray}{0.20} &
\textcolor{gray}{0.36} & \textcolor{gray}{1.10} & 
\textcolor{gray}{0.38} & \textcolor{gray}{2.21} & 
\textcolor{gray}{0.67} & \textcolor{gray}{3.67} & 
\textbf{0.20} & \textbf{0.78} & 
\textcolor{gray}{0.77} & \textcolor{gray}{5.21} &
\textbf{\textcolor{blue}{0.43}} & \textbf{\textcolor{blue}{0.86}} \\

PointGoal2 & 
\textcolor{gray}{0.44} & \textcolor{gray}{7.31} &  
\textcolor{gray}{0.53} & \textcolor{gray}{10.57} &  
\textcolor{gray}{0.29} & \textcolor{gray}{1.77} & 
\textcolor{gray}{0.30} & \textcolor{gray}{2.58} & 
\textbf{0.18} & \textbf{0.56} & 
\textcolor{gray}{0.30} & \textcolor{gray}{1.40} &
\textcolor{gray}{-0.98} & \textcolor{gray}{1.52} &
\textbf{\textcolor{blue}{0.21}} & \textbf{\textcolor{blue}{0.89}} \\

PointPush1 & 
\textcolor{gray}{0.15} & \textcolor{gray}{2.98} &  
\textcolor{gray}{0.22} & \textcolor{gray}{3.94} &  
\textbf{0.16} & \textbf{0.48} & 
\textcolor{gray}{0.21} & \textcolor{gray}{1.16} & 
\textbf{\textcolor{blue}{0.27}} & \textbf{\textcolor{blue}{0.72}} & 
\textcolor{gray}{0.14} & \textcolor{gray}{1.08} & 
\textbf{0.06} & \textbf{0.46} & 
\textbf{0.11} & \textbf{0.76} \\

PointPush2 & 
\textcolor{gray}{-0.04} & \textcolor{gray}{4.88} &
\textcolor{gray}{0.09} & \textcolor{gray}{5.32} &
\textcolor{gray}{0.13} & \textcolor{gray}{2.54} &
\textcolor{gray}{0.11} & \textcolor{gray}{3.7} & 
\textcolor{gray}{0.27} & \textcolor{gray}{1.9} &
\textcolor{gray}{0.11} & \textcolor{gray}{2.09} &
\textbf{-0.08} & \textbf{0.88} &
\textbf{\textcolor{blue}{0.04}} & \textbf{\textcolor{blue}{0.91}} \\

\cdashline{1-17}
\addlinespace[3pt]
AntVel & 
\textcolor{gray}{1.00} & \textcolor{gray}{10.29} &
\textbf{-1.01} & \textbf{0.00} & 
\textcolor{gray}{0.89} &\textcolor{gray}{3.08} &
\textbf{0.31} & \textbf{0.00} & 
\textbf{0.89} &\textbf{0.00} & 
\textbf{0.89} &\textbf{0.38}  & 
\textbf{0.32} &\textbf{0.00}  & 
\textbf{\textcolor{blue}{0.90}} & \textbf{\textcolor{blue}{0.01}} \\

HalfCheetahVel &
\textbf{0.64} & \textbf{0.00} &
\textcolor{gray}{0.08} & \textcolor{gray}{2.56} &
\textcolor{gray}{0.56} & \textcolor{gray}{1.90} & 
\textbf{0.87} & \textbf{0.23} & 
\textbf{\textcolor{blue}{0.89}} & \textbf{\textcolor{blue}{0.00}} &
\textbf{0.88} & \textbf{0.31} &
\textbf{0.85} & \textbf{0.92} &
\textbf{0.83} & \textbf{0.00} \\

HopperVel &
\textcolor{gray}{0.10} & \textcolor{gray}{3.77} & 
\textcolor{gray}{0.16} & \textcolor{gray}{13.40} &
\textbf{0.17} & \textbf{0.86} &
\textcolor{gray}{0.08} & \textcolor{gray}{7.39} &
\textbf{0.17} & \textbf{0.70} &
\textbf{0.17} & \textbf{0.01} &
\textbf{0.11} & \textbf{0.68} &
\textbf{\textcolor{blue}{0.63}} & \textbf{\textcolor{blue}{0.34}} \\

SwimmerVel & 
\textcolor{gray}{0.58} & \textcolor{gray}{23.64} &
\textcolor{gray}{0.31} & \textcolor{gray}{11.58} &
\textcolor{gray}{0.67} & \textcolor{gray}{1.47} &
\textcolor{gray}{0.42} & \textcolor{gray}{1.31} &
\textbf{-0.02} & \textbf{0.01} &
\textcolor{gray}{0.43} & \textcolor{gray}{7.01} &
\textcolor{gray}{0.00} & \textcolor{gray}{2.44} &
\textbf{\textcolor{blue}{0.47}} & \textbf{\textcolor{blue}{0.15}} \\

Walker2dVel & 
\textcolor{gray}{0.13} & \textcolor{gray}{2.63} &
\textcolor{gray}{0.23} & \textcolor{gray}{0.36} &
\textbf{0.74} & \textbf{0.32} &
\textbf{0.10} & \textbf{0.72} &
\textcolor{gray}{0.39} & \textcolor{gray}{1.10} &
\textbf{\textcolor{blue}{0.80}} & \textbf{\textcolor{blue}{0.01}} & 
\textbf{-0.01} & \textbf{0.00} &
\textbf{0.09} & \textbf{0.98} \\

\midrule
AntRun &
\textcolor{gray}{0.57} & \textcolor{gray}{1.05} & 
\textbf{0.02} & \textbf{0.00} &
\textcolor{gray}{0.72} & \textcolor{gray}{1.60} &
\textcolor{gray}{0.69} & \textcolor{gray}{2.54} &
\textbf{0.36} & \textbf{0.39} & 
\textcolor{gray}{0.52} & \textcolor{gray}{1.09} &
\textbf{{0.14}} & \textbf{{0.00}} &
\textbf{\textcolor{blue}{0.50}} & \textbf{\textcolor{blue}{0.04}} \\

BallRun &
\textcolor{gray}{0.62} & \textcolor{gray}{6.30} &
\textcolor{gray}{0.24} & \textcolor{gray}{1.90} &
\textbf{0.20} & \textbf{0.00} & 
\textcolor{gray}{0.28} & \textcolor{gray}{1.71} &
\textbf{0.18} & \textbf{0.08} &
\textbf{0.13} & \textbf{0.00} &
\textbf{{0.04}} & \textbf{{0.80}} &
\textbf{\textcolor{blue}{0.26}} & \textbf{\textcolor{blue}{0.14}} \\

CarRun & 
\textbf{0.87} & \textbf{0.00} &
\textcolor{gray}{0.98} & \textcolor{gray}{4.32} &
\textcolor{gray}{0.34} & \textcolor{gray}{1.44} & 
\textcolor{gray}{0.96} & \textcolor{gray}{1.90} & 
\textbf{0.76} & \textbf{0.00} & 
\textbf{\textcolor{blue}{0.97}} & \textbf{\textcolor{blue}{0.42}} &
\textcolor{gray}{1.68} & \textcolor{gray}{18.13} &
\textbf{\textcolor{blue}{0.97}} & \textbf{\textcolor{blue}{0.03}} \\

DroneRun & 
\textcolor{gray}{0.65} & \textcolor{gray}{7.80} &
\textcolor{gray}{0.28} & \textcolor{gray}{3.02} & 
\textbf{\textcolor{blue}{0.58}} & \textbf{\textcolor{blue}{0.32}} &
\textcolor{gray}{0.40} & \textcolor{gray}{4.99} & 
\textbf{0.17} & \textbf{0.00} &
\textcolor{gray}{0.35} & \textcolor{gray}{13.57} & 
\textcolor{gray}{0.71} & \textcolor{gray}{10.94} &
\textbf{{0.35}} & \textbf{{0.58}} \\

AntCircle & 
\textcolor{gray}{0.20} & \textcolor{gray}{7.63} &
\textbf{0.00} & \textbf{0.00} &
\textcolor{gray}{0.45} & \textcolor{gray}{4.06} &
\textcolor{gray}{0.64} & \textcolor{gray}{6.17} &
\textbf{{0.20}} & \textbf{{0.00}} &
\textbf{0.31} & \textbf{0.00} &
\textbf{\textcolor{blue}{0.52}} & \textbf{\textcolor{blue}{0.05}} &
\textbf{{0.20}} & \textbf{{0.98}} \\

BallCircle & 
\textcolor{gray}{0.70} & \textcolor{gray}{4.34} &
\textcolor{gray}{0.62} & \textcolor{gray}{1.08} &
\textbf{0.38} & \textbf{0.67} & 
\textcolor{gray}{0.67} & \textcolor{gray}{1.29} &
\textbf{0.29} & \textbf{0.00} &
\textbf{0.58} & \textbf{0.30} &
\textbf{\textcolor{blue}{0.73}} & \textbf{\textcolor{blue}{0.15}} &
\textbf{{0.49}} & \textbf{{0.81}} \\

CarCircle &
\textcolor{gray}{0.46} & \textcolor{gray}{3.53} &
\textcolor{gray}{0.71} & \textcolor{gray}{0.00} &
\textbf{\textcolor{blue}{0.72}} & \textbf{\textcolor{blue}{0.38}} &
{\textcolor{gray}{0.64}} & {\textcolor{gray}{2.11}} &
\textbf{0.39} & \textbf{0.00} &
\textbf{{0.62}} & \textbf{{0.42}} &
\textbf{\textcolor{blue}{0.72}} & \textbf{\textcolor{blue}{0.23}} &
\textbf{{0.40}} & \textbf{{0.90}} \\

DroneCircle & 
\textcolor{gray}{0.30} & \textcolor{gray}{2.23} &
\textcolor{gray}{-0.22} & \textcolor{gray}{1.83} &
\textcolor{gray}{0.51} & \textcolor{gray}{1.57} & 
{\textcolor{gray}{0.56}} & {\textcolor{gray}{4.49}} & 
\textbf{\textcolor{blue}{0.49}} &  \textbf{\textcolor{blue}{0.09}} & 
\textbf{0.45} & \textbf{0.18} & 
\textcolor{gray}{0.31} & \textcolor{gray}{1.45} &
\textcolor{gray}{-0.09} & \textcolor{gray}{4.56} \\

\midrule
easysparse & 
\textcolor{gray}{0.89} & \textcolor{gray}{9.94} &
\textbf{-0.06} & \textbf{0.10} &
\textbf{-0.03} & \textbf{0.00} &
\textcolor{gray}{0.26} &\textcolor{gray}{2.26} & 
\textbf{0.41} &\textbf{0.08} & 
\textbf{0.00} & \textbf{0.49} & 
\textbf{-0.06} & \textbf{0.16} & 
\textbf{\textcolor{blue}{0.62}} & \textbf{\textcolor{blue}{0.02}} \\

easymean &
\textcolor{gray}{0.57} & \textcolor{gray}{6.20} & 
\textbf{-0.06} & \textbf{0.10} &
\textbf{0.38} & \textbf{0.00} & 
\textbf{0.00} & \textbf{0.41} & 
\textbf{0.42} & \textbf{0.21} & 
\textbf{0.01} & \textbf{0.78} & 
\textbf{-0.06} & \textbf{0.00} & 
\textbf{\textcolor{blue}{0.60}} & \textbf{\textcolor{blue}{0.39}} \\

easydense &
\textcolor{gray}{0.47} & \textcolor{gray}{4.70} &
\textbf{-0.06} & \textbf{0.10} &
\textbf{0.50} & \textbf{0.57} & 
\textbf{0.00} & \textbf{0.28} & 
\textbf{0.48} & \textbf{0.86} & 
\textbf{0.10} & \textbf{0.49} & 
\textbf{-0.06} & \textbf{0.10} & 
\textbf{\textcolor{blue}{0.71}} & \textbf{\textcolor{blue}{0.82}} \\

mediumsparse & 
\textcolor{gray}{0.88} & \textcolor{gray}{4.97} &
\textbf{-0.06} & \textbf{0.10} &
\textbf{\textcolor{blue}{0.40}} &\textbf{\textcolor{blue}{0.11}} &
\textbf{0.10} & \textbf{0.94} & 
\textbf{{0.39}} & \textbf{{0.09}} &
\textbf{\textcolor{blue}{0.41}} & \textbf{\textcolor{blue}{0.23}} &
\textbf{-0.08} & \textbf{0.16} & 
\textbf{0.17} & \textbf{0.32} \\

mediummean &
\textcolor{gray}{0.87} & \textcolor{gray}{5.00} &
\textbf{-0.06} & \textbf{0.10} & 
\textcolor{gray}{0.63} & \textcolor{gray}{2.53} & 
\textcolor{gray}{0.19} & \textcolor{gray}{1.33} &
\textbf{\textcolor{blue}{0.59}} & \textbf{\textcolor{blue}{0.85}} &
\textcolor{gray}{0.85} & \textcolor{gray}{2.85} &
\textbf{-0.06} & \textbf{0.00} & 
\textbf{0.18} & \textbf{0.32} \\

mediumdense & 
\textcolor{gray}{0.89} & \textcolor{gray}{4.68} &
\textbf{-0.06} & \textbf{0.25} &
\textcolor{gray}{0.93} & \textcolor{gray}{4.08} &
\textbf{0.08} & \textbf{0.74} &
\textbf{\textcolor{blue}{0.50}} & \textbf{\textcolor{blue}{0.05}} &
\textcolor{gray}{0.56} & \textcolor{gray}{1.01} &
\textbf{-0.07} & \textbf{0.16} & 
\textbf{0.18} & \textbf{0.32} \\

hardsparse &
\textcolor{gray}{0.38} & \textcolor{gray}{3.56} &
\textbf{-0.05} & \textbf{0.10} &
\textbf{0.31} & \textbf{0.98} &
\textbf{0.11} & \textbf{1.11} & 
\textbf{0.32} & \textbf{0.14} & 
\textcolor{gray}{0.57} & \textcolor{gray}{1.81} & 
\textbf{-0.05} & \textbf{0.10} & 
\textbf{\textcolor{blue}{0.44}} & \textbf{\textcolor{blue}{0.32}} \\

hardmean &
\textcolor{gray}{0.33} & \textcolor{gray}{3.59} & 
\textbf{-0.05} & \textbf{0.16} &
\textbf{0.09} & \textbf{0.01} & 
\textbf{-0.05} & \textbf{0.09} &
\textbf{0.25} & \textbf{0.36} & 
\textbf{0.17} & \textbf{0.16} & 
\textbf{-0.05} & \textbf{0.10} & 
\textbf{\textcolor{blue}{0.45}} & \textbf{\textcolor{blue}{0.06}} \\

harddense & 
\textcolor{gray}{0.23} & \textcolor{gray}{2.50} &
\textbf{-0.04} & \textbf{0.10} &
\textbf{0.23} & \textbf{0.03} &
\textbf{0.02} & \textbf{0.14} &
\textbf{0.30} & \textbf{0.47} & 
\textcolor{gray}{0.36} & \textcolor{gray}{1.85} &
\textbf{-0.04} & \textbf{0.10} &
\textbf{\textcolor{blue}{0.43}} & \textbf{\textcolor{blue}{0.45}} \\

\midrule
\textbf{Safe / Optimal}  & 2  & 0  & 14  & 1  & 16  & 4  & 15  & 2  & 27  & 7  & 21  & 4  &20  & 3 &36  & 21 \\

\textbf{Average} & \textcolor{gray}{0.46} & \textcolor{gray}{6.72} &\textcolor{gray}{0.20} & \textcolor{gray}{7.54}&\textcolor{gray}{0.41} & \textcolor{gray}{2.61}&\textcolor{gray}{0.28} & \textcolor{gray}{1.94}&\textcolor{gray}{0.35} & \textcolor{gray}{1.90}& \textcolor{gray}{0.35} & \textcolor{gray}{1.66} & \textcolor{gray}{0.18} & \textcolor{gray}{5.38} & \textbf{\textcolor{blue}{0.33}} & \textbf{\textcolor{blue}{0.70}}\\
\bottomrule
\end{tabular}
}
\end{center}
\vskip -0.2in
\end{table*}

In this section, we evaluate SDGD on the DSRL benchmark \cite{liu2023datasets}, which includes 38 tasks from Safety-Gymnasium \cite{ji2023safety}, Bullet-Safety-Gym \cite{Gronauer2022BulletSafetyGym}, and MetaDrive \cite{li2022metadrive}. 
We focus on four questions: 
(1) whether SDGD improves the reward--cost trade-off under a strict safety threshold?
(2) whether it adapts to different deployment-time cost limits? 
(3) whether it handles cost limits that change within an episode? and 
(4) how each guidance component contributes to safety and performance?
SDGD achieves a decision efficiency of 2.1 Hz on an NVIDIA A6000 GPU, faster than prior methods like TREBI (1.1 Hz) and HD (1.0 Hz).

\begin{figure*}
\vskip -0.1in
\begin{center}
\centerline{\includegraphics[width=\textwidth]{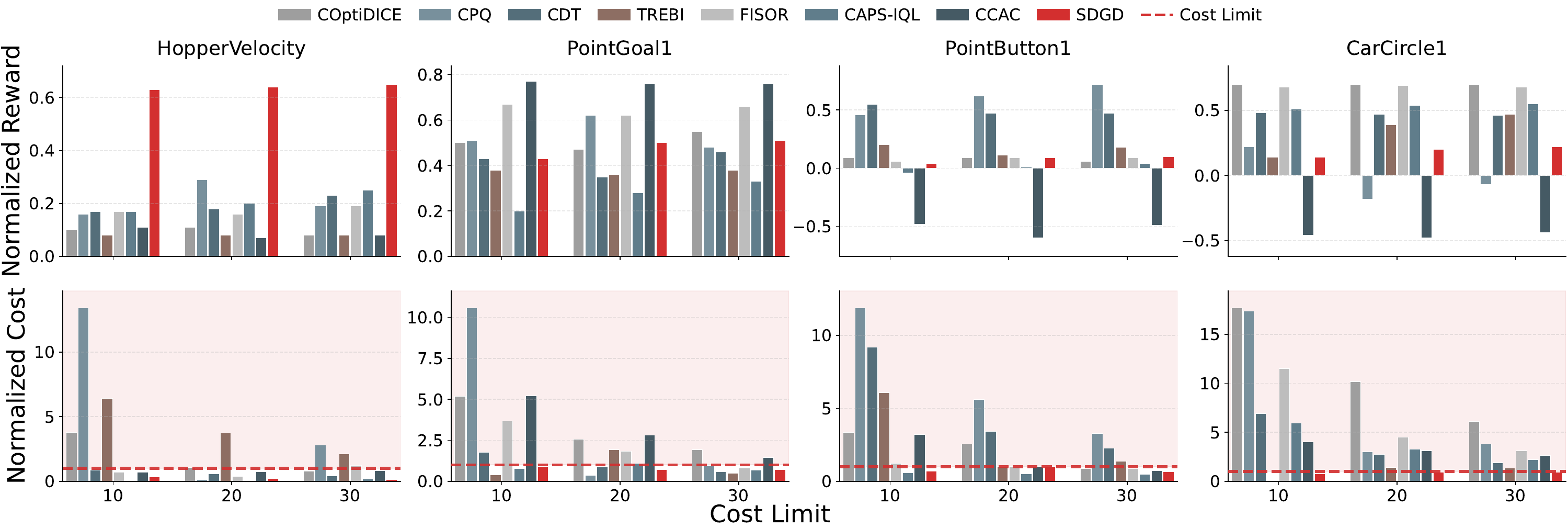}}
\vskip -0.13in
\caption{
Performance comparison across different cost limits (10, 20, 30) in four tasks.
}
\label{fig: real-time budgets}
\end{center}
\vskip -0.25in
\end{figure*}

\textbf{Evaluation Setup.}
We report normalized reward and normalized cost following the DSRL normalization protocol \cite{liu2023datasets}. 
A method is considered safe on a task if its normalized cost satisfies \(C_{\mathrm{normalized}}\le 1\). 
For a unified and stricter comparison, we evaluate all methods under an absolute cost limit of 10, rather than using the larger thresholds in the original DSRL protocol (e.g., \{10, 20, 40\} for BulletGym and MetaDrive, and \{20, 40, 80\} for SafetyGym). 


\begin{figure*}[b]
\vskip -0.3in
\begin{center}
\centerline{\includegraphics[width=\textwidth]{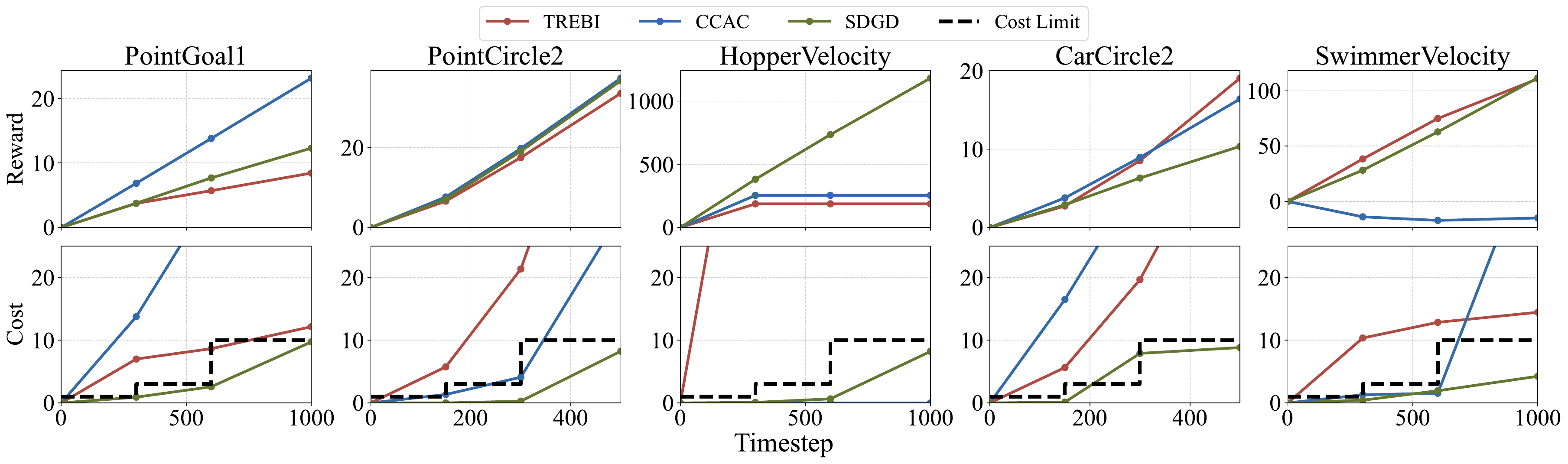}}
\vskip -0.1in
\caption{
Performance under dynamically time-varying cost limits. The experiment evaluates the adaptability of TREBI, CCAC, and SDGD by varying the cost limit within a single episode (from 1, to 3, to 10). 
The dashed line represents the changing cost limit. 
}
\setlength{\abovecaptionskip}{20pt}
\label{fig:dynamic_limits}
\end{center}
\vskip -0.2in
\end{figure*}

\begin{table}
    \centering
    \small
    \vspace{-5pt}

    \begin{minipage}{0.48\textwidth}
        \centering
        \caption{Ablation study on guidance mechanism. SDGD outperforms the baseline in safety.}
        \label{tab:swap_ablation}
        \vspace{-2pt}
        \setlength{\tabcolsep}{2pt}
        \begin{tabularx}{\linewidth}{lCCCC} 
            \toprule
            {Baseline} & \multicolumn{2}{c}{Switch} & \multicolumn{2}{c}{SDGD} \\
            \cmidrule(lr){2-3} \cmidrule(lr){4-5}
            {Task} & {R $\uparrow$} & {C $\downarrow$} & {R $\uparrow$} & {C $\downarrow$} \\
            \midrule
            CarButton1 & \textcolor{gray}{0.06} & \textcolor{gray}{2.01} & \textbf{-0.04} & \textbf{0.63} \\
            CarButton2 & \textcolor{gray}{0.02} & \textcolor{gray}{1.11} & \textbf{-0.05} & \textbf{0.62} \\
            HopperVel  & \textcolor{gray}{0.18} & \textcolor{gray}{1.92} & \textbf{0.63} & \textbf{0.34} \\
            CarRun     & \textcolor{gray}{2.18} & \textcolor{gray}{3.84} & \textbf{0.97} & \textbf{0.03} \\
            DroneRun   & \textcolor{gray}{0.77} & \textcolor{gray}{9.38} & \textbf{0.35} & \textbf{0.58} \\
            \bottomrule
        \end{tabularx}
    \end{minipage}
    \hfill
    \begin{minipage}{0.48\textwidth}
        \centering
        \caption{Stability analysis of CFG weight $w$ and CG scale $\lambda$ on MetaDrive. {R}: Reward, {C}: Cost. \colorbox{gray!20}{Gray} cells indicate unsafe results.}
        \label{tab:stability}
        \vspace{-3pt}
        \setlength{\tabcolsep}{1.2pt} 
        \begin{tabularx}{\linewidth}{c*{8}{C}} 
            \toprule
            $\lambda$ & \multicolumn{2}{c}{0.01} & \multicolumn{2}{c}{0.02} & \multicolumn{2}{c}{0.04} & \multicolumn{2}{c}{0.08} \\ 
            \cmidrule(lr){2-3} \cmidrule(lr){4-5} \cmidrule(lr){6-7} \cmidrule(lr){8-9}
            $w$ & R & C & R & C & R & C & R & C \\
            \midrule
            1 & 0.39 & 0.14 & 0.42 & 0.39 & 0.49 & 0.78 & \cellcolor{gray!20}0.51 & \cellcolor{gray!20}1.84 \\
            2 & 0.40 & 0.13 & 0.42 & 0.36 & 0.44 & 0.59 & 0.46 & 0.87 \\
            4 & 0.41 & 0.11 & 0.43 & 0.30 & 0.42 & 0.41 & 0.45 & 0.52 \\
            8 & 0.41 & 0.11 & 0.43 & 0.16 & 0.35 & 0.18 & 0.31 & 0.21 \\
            \bottomrule
        \end{tabularx}
    \end{minipage}

    \vspace{-20pt}
\end{table}

\textbf{Baselines.}
We compare with seven baselines:
\textit{COptiDICE} \cite{lee2022coptidice}, a stationary distribution correction method based on OptiDICE \cite{lee2021optidice};
\textit{CPQ} \cite{xu2022constraints}, a penalty-based method for out-of-distribution actions;
\textit{CDT} \cite{liu2023constrained}, a Decision Transformer method with safety constraints;
\textit{TREBI} \cite{lin2023safe}, a diffusion-based planner using classifier guidance for safe trajectory generation;
\textit{FISOR} \cite{zheng2024safe}, a feasibility-guided offline RL method;
\textit{CAPS} \cite{chemingui2025constraint}, which adapts to cost limits by switching among policies; and
\textit{CCAC} \cite{guo2025constraint}, an actor-critic method that conditions the actor and critics on the cost limit.

\begin{figure}
\vskip -0.1in
\begin{center}
\centerline{\includegraphics[width=\columnwidth]{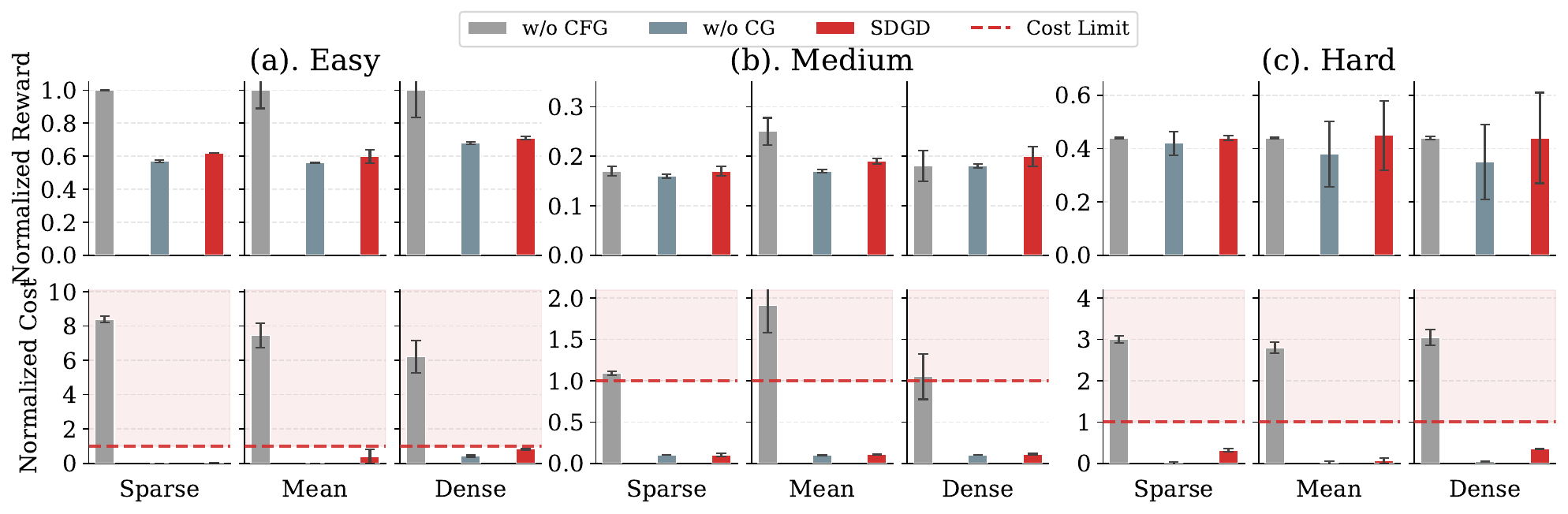}}
\vskip -0.1in
\caption{
Ablation study on the guidance components. The SDGD is compared against two variants: one without Classifier Guidance (w/o CG) and another without Classifier-Free Guidance (w/o CFG).
}
\setlength{\abovecaptionskip}{20pt}
\label{fig:decoupled_guidance_ablation}
\end{center}
\vskip -0.1in
\end{figure}

\textbf{Main Results.}
Table~\ref{tab: main_result1} shows that SDGD achieves the strongest safety compliance among compared methods, satisfying the normalized cost constraint on 36 out of 38 tasks. 
Among methods that satisfy the constraint, SDGD obtains the highest reward on 21 tasks, indicating a favorable reward--cost trade-off under the strict evaluation threshold. 
Adaptive actor-critic baselines, CAPS and CCAC, satisfy the constraint on 21 and 20 tasks, respectively, but achieve the best safe reward on fewer tasks. 
Generative baselines such as CDT, TREBI, and FISOR improve over earlier approaches in safety compliance, but still fail on several tasks where SDGD remains safe. 
Earlier methods show complementary limitations: CPQ is often conservative with lower reward, whereas COptiDICE frequently violates the cost constraint. 


\subsection{Ablation Study and Analysis}

\textbf{Ablation on Cost Limits.}
To evaluate adaptation to different safety budgets, we compare SDGD with baselines under cost limits of 10, 20, and 30 in Figure~\ref{fig: real-time budgets}. 
SDGD maintains normalized cost below the specified limit while improving reward as the budget becomes less restrictive. 
Most baselines either violate the constraint at tighter budgets or remain overly conservative at looser budgets. 
All results use a single trained SDGD model; no retraining is performed when the cost limit changes.


\textbf{Time-Varying Limits.}
We next evaluate whether a single model can respond to cost limits that change within an episode. 
As shown in Figure~\ref{fig:dynamic_limits}, the limit is varied from 1 to 3 to 10 during evaluation. 
SDGD adjusts its behavior as the limit changes, keeping cost below the active threshold while improving reward when the budget becomes less restrictive. 
This suggests that cost-limit conditioning can support dynamic budget adaptation without retraining.


\textbf{Decoupled Guidance Components.}
Figure~\ref{fig:decoupled_guidance_ablation} evaluates the two guidance components by removing reward-gradient guidance (w/o CG) or cost-limit conditioning through classifier-free guidance (w/o CFG).
Removing reward-gradient guidance reduces return, while removing cost-limit conditioning leads to cost violations, showing that the two mechanisms play complementary roles.



\textbf{Swapping Guidance Roles.}
To test whether guidance roles are interchangeable, we evaluate an inverted `CG-Cost/CFG-Reward' baseline (Table~\ref{tab:swap_ablation}).
This variant violates safety constraints on all representative tasks, suggesting that CG is poorly suited for strict safety limits.

\begin{figure}
    \centering
    \vskip -0.2in
    \includegraphics[width=\columnwidth]{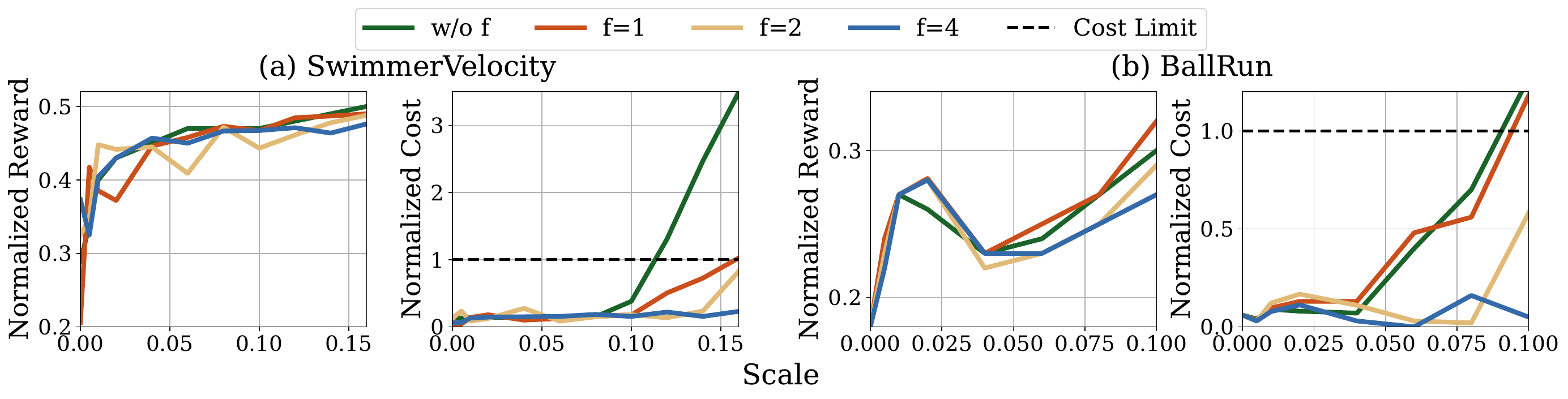}
    \vskip -0.2in
    \caption{Ablation on the reward guidance scale and Feasible Length $f$ for trajectory re-labeling. 
    Without re-labeling ('w/o $f$'), stronger guidance boosts rewards but violates the cost limit, whereas re-labeling ($f>0$) maintains safety, with larger $f$ values encouraging conservative behavior.
    }
    \vskip -0.2in
    \label{fig: Hyperparameter}
\end{figure}

\textbf{Hyperparameter Choices.}
Figure \ref{fig: Hyperparameter} confirms that trajectory re-labeling (Feasible Length $f$) is essential for preventing safety violations under strong reward guidance.
Furthermore, the stability analysis in Table \ref{tab:stability} demonstrates SDGD's robustness: it maintains strict safety compliance (Normalized Cost $\le 1$) across a broad range of CFG weights $w$ and CG scales $\lambda$.


\section{Conclusion and Future Work}

We presented \textit{Safe Decoupled Guidance Diffusion} (SDGD), a diffusion-based planner that decouples cost limit conditioning from reward-gradient guidance for adaptive safe offline reinforcement learning. 
With Feasible Trajectory Relabeling, SDGD mitigates cost increases induced by reward optimization while preserving return improvement. 
Experiments on DSRL \cite{liu2023datasets} show strong safety compliance, high reward among methods satisfying the constraint, and adaptation to changing cost limits without retraining. 
Future work will study faster sampling and real-world validation.


\bibliographystyle{unsrtnat}
\bibliography{references}

\end{document}